\RequirePackage{fix-cm}
\documentclass[twocolumn]{svjour3}          
\smartqed  
\usepackage{graphicx}
\usepackage{subfigure}
\usepackage{color}

\usepackage{multirow}
\begin{document}

\title{Multi-task, multi-label and multi-domain learning with residual convolutional networks for emotion recognition}


\author{Gerard Pons         \and
        David Masip 
}


\institute{G. Pons and D.Masip \at
              Department of Computer Science, Universitat Oberta de Catalunya, Spain \\
              \email{\{gponsro, dmasipr\}@uoc.edu}           
}

\date{Received: date / Accepted: date}

\maketitle

\begin{abstract}
Automated emotion recognition \emph{in the wild} from facial images remains a challenging problem. Although recent advances in Deep Learning have supposed a significant breakthrough in this topic, strong changes in pose, orientation and point of view severely harm current approaches. In addition, the acquisition of labeled datasets is costly, and current state-of-the-art deep learning algorithms cannot model all the aforementioned difficulties. In this paper, we propose to apply a multi-task learning loss function to share a common feature representation with other related tasks. Particularly we show that emotion recognition benefits from jointly learning a model with a detector of facial Action Units (collective muscle movements). The proposed loss function addresses the problem of learning multiple tasks with heterogeneously labeled data, improving previous multi-task approaches. We validate the proposal using two datasets acquired in non controlled environments, and an application to predict compound facial emotion expressions. 


\keywords{Facial emotion recognition \and Facial Action Units \and Multi-task learning \and Convolutional neural networks}
\end{abstract}

\section{Introduction}
\label{sec:intro}

Images of faces provide relevant information for emotion perception. As humans we can infer an accurate first impression of somebody's emotions by observing their face. Multiple applications benefit from automated facial emotion recognition, such as human computer interaction (HCI)~\cite{bartlett2003real}, student engagement estimation~\cite{whitehill2014faces}, emotionally aware devices~\cite{soleymani2013emotionally}, or the improvement of expression production in autism disorder patients~\cite{cockburn2008smilemaze}.

The state-of-the-art in automated facial expression analysis shows excellent performance in the controlled scenario, where images are acquired in studio environments. Nevertheless, the categorization of emotions \textsl{in the wild}, is still an unsolved problem. Besides the strong intra-class variability, facial expression algorithms \textsl{in the wild} must also deal with strong local changes in the illumination conditions, out of plane rotations, large variations in pose and point of view, and low resolution imaging. 

Recent advances in computer vision and particularly in object recognition suggest that new methods based on Deep Learning can improve the facial expression recognition task. Convolutional Neural Networks (CNNs) have represented a relevant breakthrough, especially since the last improvements on the ImageNet Challenge \cite{Krizhevsky2012}. However, the amount of available data for this task is small, especially in all the possible configurations of pose, illumination and resolution. This supposes an inconvenience to exploit the training capacity of these networks, which need large amounts of training data. In this context the introduction of multi-task learning is particularly relevant, as it proved to successfully boost the performance of an individual task with the inclusion of other correlated tasks in the training process~\cite{Ganin_2015,Hinton_2015_MT,Zhu_2012}. Thus, tasks with small amounts of data available can benefit from being trained simultaneously with other tasks, sharing a common feature representation and transferring knowledge between different domains. One of the main difficulties with multi-task approaches using different databases is the fact that not all the samples are labeled for all the tasks. In order to deal with this, classical approaches define different loss functions for each task and train alternatively for the different domains. This penalizes the tasks where sample labels are not available. One approach to solve this problem is a selective joint loss, but it only predicts probabilities for the label set to which the image belongs. Hence, in order to generate predictions for all the tasks, in this paper we propose a multi-label database-wise joint loss to overcome this problem.




This paper makes the following contributions: (i) we formalize a novel dataset-wise selective sigmoid cross-entropy loss function to simultaneously train a multi-task, multi-label and multi-domain model. (ii) We validate that this new proposal outperforms single task CNNs and the classical multi-task approach using different databases. (iii) We show that the results of the joint learning of an unlabeled task are coherently correlated to the labeled task in the case of the emotion recognition problem.

\section{Related work}
\label{sec:rel_work}

Automated emotion recognition methods from facial expression analysis in images use a subset of discrete basic emotions, as defined in~\cite{ekman1971constants}. Although the study of emotional states from faces is grounded in the late 19th century~\cite{darwin1872expression}, Ekman and Friesen~\cite{ekman1971constants} defined a set of six basic emotions that are shared among all cultures, namely \textsl{happiness}, \textsl{surprise}, \textsl{anger}, \textsl{sadness}, \textsl{fear}, \textsl{disgust}. The production of these emotions in faces depends on specific facial muscle movements. The Facial Action Coding System (FACS)~\cite{ekman78facs} defines a set of Action Units (AUs) that atomically group muscle movements. Each expression of emotion can be encoded as a combination of AU activations.

Facial expression recognition is a multidisciplinary research field, studied in machine learning, computer vision, cognitive science, psychology, neuroscience and applied health sciences. The number of computer vision researchers working in the field of facial expression analysis has increased since the early 90s, and a large amount of published works in the topic exist. Depending on the features used for the recognition task, we can distinguish two prevailing methodologies: geometric based approaches and appearance based approaches. In the first case, algorithms focus on localizing and tracking specific fiducial facial landmarks, in order to train a classifier based on distances and relative positions of these landmarks. Kotsia and Pitas~\cite{kotsia2007facial} track a set of 119 key points to classify emotions from sequences, and Jeni et al.~\cite{jeni2011high} apply a Procrustes transformation on a reduced subset of 117 landmarks for emotion recognition. In appearance based emotion recognition, a set of features is extracted from pixel images to train a classifier. Classic examples are Gabor Wavelets~\cite{bartlett2003real}, Histogram of Oriented Gradients (HOG)~\cite{li2009facial}, and Local Binary Patterns (LBP)~\cite{zhao2007dynamic}. More complex models use Gaussian processes~\cite{eleftheriadis2015discriminative}, hybrid methods using mid-level features~\cite{sanchez2015emotion} or patches from specific key locations~\cite{happy2015automatic}. More recently, deep learning methods have outperformed most state-of-the-art algorithms in the literature. Liu et al.~\cite{liu2014deeply} apply deep learning to a geometric model of facial regions for facial expression analysis, and Lu et al.~\cite{lu2016convolutional} use Convolutional Neural Networks (CNN) on facial appearance. Recent models obtain improved results focusing on facial expression recognition \textsl{in the wild} by using committees of several CNN classifiers~\cite{kim2015hierarchical,pons2017supervised}.
Although most of the previous models focus on static images, facial expression analysis can benefit from temporal information. Cohen et al.~\cite{cohen2003facial} use Hidden Markov models on video sequences. Deep learning methods for emotion perception modeling in video include the use of CNNs for feature extraction and LSTM for learning the temporal dynamics~\cite{sun2016lstm}. Recent surveys on facial expression analysis can be found at~\cite{corneanu2016survey,martinez2017automatic,sariyanidi2015automatic}.

Despite the current improvements, emotion recognition in the wild remains an open problem for the computer vision community. Results from the Emotion Recognition in the Wild (EmotiW 2015) reveal modest accuracies~\cite{dhall2015video}. EmotionNet is the biggest Challenge in terms of data available, with more than 1 million images (2,000 labeled emotions and 950K unlabeled samples). The first edition of the challenge concluded that non-frontal faces still pose major difficulties to automated algorithms, and recognition rates decrease as a function of pitch and yaw rotations~\cite{benitez2017emotionet}.

Deep learning methods benefit from large training datasets, but the labeling process of facial expression data requires expert FACs coders and it is a costly task. Therefore, most of the current available data sets are reduced and difficult to acquire. In this paper we conjecture that emotion recognition can be improved if we transfer knowledge from other related tasks. Particularly we focus on Multi-task learning (MTL). MTL was first studied by Caruana in~\cite{caruana1998multitask}, where he proposes to jointly learn parallel tasks sharing a common representation, and transferring part of the knowledge learned to solve one task to improve the learning of the other related tasks. Detailed analysis on the topic is carried out in some recent surveys~\cite{zhang2017survey,pan2010survey}.

Ranjan et al.~\cite{ranjan2017hyperface} propose a multi-task approach for face detection, landmark localization, pose estimation and gender recognition. This method exploits the benefits of multi-task to improve the performance of each individual tasks. However, they use a single source of data, hence all the images must be labeled with all the tasks involved.

The closest work to this paper was published by Fourure et al.~\cite{fourure2017multi}. They propose a multi-task CNN  for semantic segmentation of outdoor images in order to learn different tasks using images from different databases. They use a selective soft-max cross entropy with the objective of not penalizing the training of a task when feeding the model with images from another task. This approach though does not learn from similar labels. For instance, when training a database A with a class labeled \emph{Grass}, and database B with a class labeled \emph{Vegetation}, the method would predict the label corresponding to the database where the input image belonged to (if the image is from database A will predict it as Grass). In this paper we hypothesize that general accuracies will improve if we learn from both labels at the same time, and take full benefit of training related tasks using different non-homogeneous datasets.


\section{Proposed approach}
\label{sec:proposal}

We propose a novel multi-label loss function that can be integrated into convolutional neural networks to improve the training of the emotion recognition task incorporating complementary tasks and data from different sources. This proposal can be extended to other multi-label problems, but in this article we focus on the multi-task training of emotion recognition and facial Action Units (AUs). 

Emotion recognition is a problem where the model classifies the input image as one of the 7 emotion classes ~\cite{ekman1971constants} (including the neutral class). Therefore, when addressing this task using a CNN, the computation of the output probabilities is performed by using soft-max function in the values of the last layer. This setting is not valid in our multi-task approach, since AU recognition is a multi-label problem. The classical setting for multi-label approaches is the use of the sigmoid function for the calculation of the output probabilities. However, using images from different sources and for different tasks implies having partially unlabeled data when performing the multi-task training, i.e. not all data is labeled with respect to all the tasks. Thus, this approach is not valid when dealing with images from different databases, since the sigmoid function penalizes the absence of labels in the related task. For example, optimizing the \emph{Happy} class using samples with no AU annotations would penalize the AU6 and AU12 classes provoking a non-desired effect since these AUs together form the emotion \emph{Happy}. We thus propose the \emph{Selective Joint Multi-task} (SJMT) approach which defines a novel dataset-wise selective sigmoid cross-entropy loss function to address multi-task, multi-label and multi-domain problems.

Given a set of images from $k$ different datasets, where the label spaces are different, we define $y_j^k$ the actual label of the $j$-th class in the space $\mathcal{L}^k$. The aim is to learn a nonlinear mapping presented as a CNN which minimizes a cross-entropy loss function for each individual sample. This function is defined as follows: 


\begin{equation}
    E(\hat{y},y,k) = \frac{-1}{N}\sum_{j \in \mathcal{L}^k}  [y_j\log{\hat{y_j}}+(1-\hat{y_j})\log{(1-\hat{y_j})}]
    \label{eq:eq1}
\end{equation}

\noindent where $N$ is the number of classes, $k$ is the number of different datasets, each defined on its own label set $\mathcal{L}^k$. The term $y_j$ is the actual label of the $j$-th class, and $\hat{y_j}$ is a sigmoid function defined as: 

\begin{equation}
\centering
	\hat{y_j} = \frac{1}{1+e^{-\hat{p}_j}}
\end{equation}

\noindent where $\hat{p}_j$ is the output of the last layer of the CNN defined as $f(Wh+b)$, $f$ is the activation function of the layer, $W$ and $b$ the weights and bias of the layer and $h$ the hidden representation of the last layer.

\begin{figure*}[ht]
  \centering
  \subfigure[]{\label{fig:esquemes_a}\includegraphics[width=60mm]{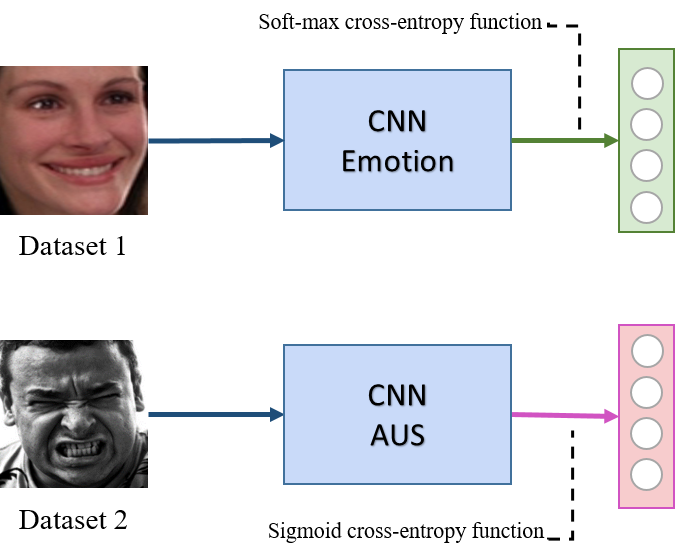}} \qquad
  \subfigure[]{\label{fig:esquemes_b}\includegraphics[width=60mm]{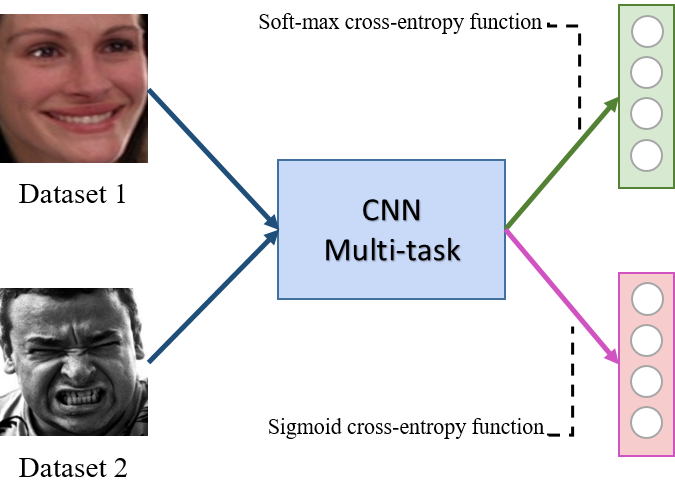}}
  \subfigure[]{\label{fig:esquemes_c}\includegraphics[width=60mm]{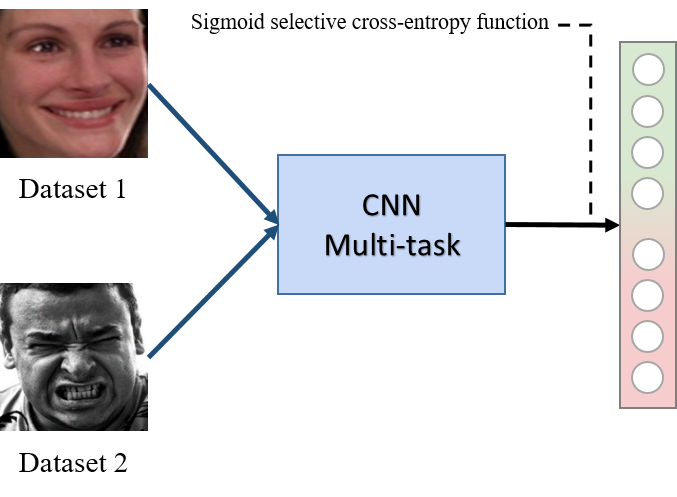}}
  \caption{Different CNN strategies used in this paper. (a) shows the traditional approach with different CNNs for different tasks. (b) shows the classical multi-task approach with different output layers and loss functions for each individual task. (c) shows the proposed multi-task approach with a single output layer and the selective sigmoid cross entropy function.}
  \label{fig:esquemes}
\end{figure*}

Figure~\ref{fig:esquemes} shows the different approaches compared in this article. First Figure~\ref{fig:esquemes_a} shows the approach where individual CNN are tailored to each task and database. Second Figure~\ref{fig:esquemes_b} shows the classical multi-task approach where the filters of the network are shared during training among the different tasks, taking into account that each task has an specific cross-entropy loss function and output layer, and the training is performed alternatively feeding the model with batches of images from each dataset. Finally, Figure~\ref{fig:esquemes_c} shows the approach proposed in this paper, where the whole network is shared among the different tasks and images from different databases are feed indistinctly thanks to the selective cross-entropy function defined in equation~\ref{eq:eq1}.  

Even though this cross-entropy function could be incorporated in any CNN architecture, in this paper we used a ResNet~\cite{He2015} to implement the multi-task, multi-label and multi-database approach. Residual networks have shown great robustness in different fields~\cite{Feichtenhofer2016_nips,Pohlen2017_semseg,Yu2017_melanoma} given their framework which eases the training of these networks, and enables them to be substantially deeper and, thus, obtain better performances than other networks.

\section{Experimental results}
\label{sec:experiments}

In this paper we used the following databases for the experimental validation:
\begin{itemize}
\item \textbf{SFEW2.0 database}~\cite{dhall2015video}, released for a competition in the 3rd Emotion Recognition In the Wild 2015 (EmotiW2015) challenge. The database was created by extracting frames from film clips with emotional content in order to obtain images in close-to-real world conditions. The images are labeled with 7 expressions (angry, disgust, fear, happy, sad, surprise, and neutral). The database consists of 958 images for training, 436 for validation, and 372 for testing. Since the labels of the test dataset are not provided, the validation set was used as test set for the experiments in this work. The database also provides the same data after being processed with a face alignment algorithm. We used this set in the experiments of this paper.

\item \textbf{EmotioNet database}~\cite{EmotioNet}, released for the EmotioNet Challenge 2017~~\cite{benitez2017emotionet}. The database consists of 25,000 images of facial expressions with manual AU annotations. These annotations were given by expert human coders and cross referenced for verification of accuracy. From these images, a subset of 2,000 images were annotated with the 7 emotion expressions and 9 compound emotions made by the combination of two basic emotions. Since the test dataset is private and was only available for participants of the challenge, the test set for the experiments in this work was generated randomly selecting 1,000 images (the same in all the experiments).

\item \textbf{Extended Cohn-Kanade database (CK+)}~\cite{CK}, consists of 593  
posed video sequences recorded from 123 university students ranging from 18 to 30 years old. The subjects were asked to express a series of facial displays including single or combined action units. The database also provides labels for the presence of emotions in these sequences. In this paper, we randomly generated a subset of 100 images for testing the methods. This database was acquired in controlled environments, and it has been used only to determine the parameters used for the neural networks. Theses paremeters were consistent in all the experiments. 

\end{itemize}

For ResNet~\cite{He2015} we used a training batch size of 128 images, and learning rate of 0.05 and exponential decay every 10,000 steps, for a total of 80,000 steps. Batch normalization and data augmentation by means of padding and random cropping images during training were used to prevent overfitting. Input data of size $32 \times 32$ was used to pre-load CIFAR10 weights~\cite{He2016}. In the comparison among individual networks we also used  VGG~\cite{Simonyan14c}. For this network we used a training batch size of 32 images, and an exponentially decayed learning rate of 0.00001 during 2,000 steps. To prevent the network from overfitting a dropout rate of 0.5 was used. In order to pre-load the trained weights used by the VGG team in the ILSVRC-2014 competition, input images of size $224 \times 224$ were used.

All the networks trained in this paper were pre-trained using the Facial Expression Recognition (FER) 2013 database~\cite{FERdatabase}, which consists of 28,709 examples for training, 3,589 for validation and 3,589 for testing. All the experiments in this work were developed using TensorFlow~\cite{tensorflow2015} on NVIDIA GeForce GTX Titan GPU.

With regard to the detection of AUs and recognition of emotion categories, the evaluation criteria used was accuracy. Accuracy measures the number correctly classified examples, indicating whether our algorithm is able to discriminate between sample images with a certain AU/emotion present. In statistics, this difference between the measured and true value is generally called observational error, and it is defined as follows: 

\begin{equation}
	accuracy_i = \frac{TP_i+TN_i}{N}
\end{equation}

\noindent where $i$ specifies the class, i.e., AU$i$ or the $i$-th emotion category, $TP_i$ (true positives) are correctly identified test instances of class $i$, $TN_i$ (true negatives) are test images correctly labeled as not belonging to class $i$, and $N$ is the total number of test images.

\subsection{Emotion and AU recognition with individual networks}

Before validating the proposed approach in this paper we needed to validate the performance of each task trained with dedicated individual networks for such task and database. In addition, the baseline accuracy for each database is reported if known.

Thus, we trained specific networks for emotion recognition (see Figure~\ref{fig:esquemes_a}) with the SFEW and CK+ databases, and for AU detection with the EmotioNet and CK+ databases. Given their outstanding results in different fields, we decided to compare two of the most used architectures in image processing, VGG-16 and ResNet (with different sizes: 32 and 110). In this comparison we also added the results of other state-of-the art commercial methods, such as the Microsoft Azure emotion recognition API~\cite{Copeland2015_azure} and the OpenFace framework for AU detection~\cite{Openface2015}. Note that these methods were not validated with the CK+ database, since we used this database only for defining the hyper-parameters used in the networks we needed to train (VGG and ResNets).


\begin{table}[ht]
	\renewcommand{\arraystretch}{1.3}
	\caption{Results of different methods for the emotion and AU recognition tasks in percentage of accuracy.}
	\label{tab:tab_results_individual}
	\centering
	\begin{tabular}{*5c}	
		\hline
        						& \multicolumn{2}{c}{Emotion} & \multicolumn{2}{c}{AUs} \\
		Method 					&  SFEW  & CK+ & EmotioNet & CK+ \\
		\hline
		Baseline		    	& 35.9\%  & - & - & 94.5\% \\
        OpenFace				& -	& - & 87.7\% & - \\
        Azure					& 33.8\% & - & - & - \\
		VGG-16			   		& 37.5\%  & 93.0\%  & 93.2\% & 95.0\%\\
        ResNet-32				& 37.8\%  & 98.0\%  & 93.2\% & 99.0\% \\
        ResNet-110				& \textbf{41.3\%}  & \textbf{99.0\%} & \textbf{93.9\%} & \textbf{99.1\%} \\
		\hline
	\end{tabular}
\end{table}

Table~\ref{tab:tab_results_individual} shows the accuracy obtained for each CNN for the different tasks and databases. Note that the images were not pre-processed specifically to enhance the performance in one database or another, and the same parameters were used in all the databases for a fare comparison. Hence, the most noticeable fact are the results obtained by ResNet-110, which are superior to the rest of the methods in the comparison.

\subsection{Multi-task learning for emotion and AU recognition}

In this section we want to validate the hypothesis that the proposed SJMT method (see Figure~\ref{fig:esquemes_c}) outperforms the individual networks compared in the previous section as well as the classical multi-task approach (see Figure~\ref{fig:esquemes_b}). Given the results from the individual network comparison we decided to use the two best architectures in this comparison: ResNet 32 and 110. 

For this experiment we jointly trained the multi-task approaches for emotion and AU recognition with the SFEW and EmotioNet databases respectively and the obtained results are summarized in Table~\ref{tab:tab_results_multi}. As occurred in the previous experiment, ResNet 110 generally obtained better results than ResNet 32. Comparing both the SJMT and classical multi-task approaches, the proposed method outperforms the classical approach in all the experiments except for AU recognition with ResNet 32. Finally, SJMT with ResNet 110 obtained the best results for both emotion and AU recognition, compared to the classical multi-task, and the single task approaches shown in Table~\ref{tab:tab_results_individual}

\begin{table}[ht]
	\renewcommand{\arraystretch}{1.3}
	\caption{Results of multi-task learning for the emotion and AU recognition tasks in percentage of accuracy. \emph{Multi-task} refers to the classical approach (see Figure~\ref{fig:esquemes_b}) and \emph{SJMT} refers to the \emph{Selective Joint Multi-task} (SJMT) approach (see Figure~\ref{fig:esquemes_c})}
	\label{tab:tab_results_multi}
	\centering
	\begin{tabular}{*5c}	
		\hline
		\multirow{2}{*}{Method} 		& Network & SFEW  & EmotioNet \\
        								&	& (Emotion) & (AUs) \\
		\hline
		\multirow{2}{*}{Multi-task}  	& ResNet-32  &  39.4\% & 93.7\% \\
						   				& ResNet-110  &  40.3\% & 93.6\% \\
		\hline
		\multirow{2}{*}{SJMT}			& ResNet-32  &  40.6\% & 93.5\% \\
        								& ResNet-110  & \textbf{45.9\%} & \textbf{93.9\%}\\

		\hline
	\end{tabular}
\end{table}

\begin{figure}[!ht]
	\centering
	\includegraphics[width=0.5\textwidth]{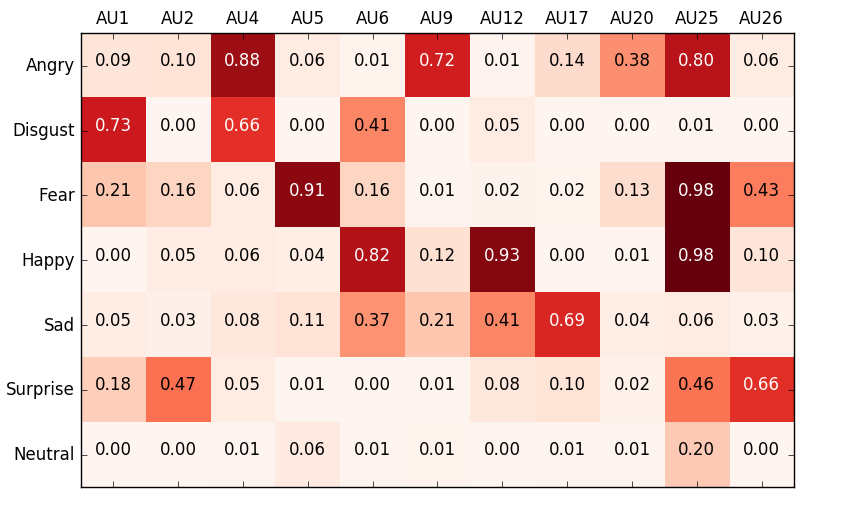}
	\caption{Mean scores for AUs in SFEW images where only emotion labels are provided using the SJMT approach.}
	\label{fig:AUs_map}
\end{figure}

In order to analyze the benefits of using the proposed SJMT approach with related tasks, we focused on the predictions for AUs when feeding the model with images from the SFEW database, which only provides labels for emotion recognition. Thus, we grouped the images by the predicted emotion category and computed the mean scores for each AU as shown in Figure~\ref{fig:AUs_map}. This figure shows that when the model recognizes an emotion category, is simultaneously capable to identify the corresponding AUs in most of the cases. In order to simplify the interpretation of the figure, Table~\ref{tab:tab_desc_aus} shows the description of the AUs predicted by SJMT along with each emotion and a representative image from the test dataset of the SFEW database.



\begin{table}[ht]
	\renewcommand{\arraystretch}{1.3}
	\caption{Description of the recognized AUs for each emotion.}
	\label{tab:tab_desc_aus}
	\centering
	\begin{tabular}{*4c}	
		\hline
		Emotion		& AU & Description  & Example \\
        								
		\hline
		\multirow{3}{*}{Angry}  	& 4  &  Brow Lowerer & \multirow{3}{*}{\includegraphics[width=0.1\textwidth]{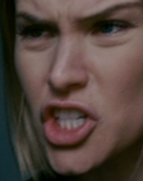}} \\
						   			& 9  &  Nose Wrinkler &  \\
                                    & 25  &  Lips Part &  \\
        \\
        \\ 
		\hline
        \multirow{3}{*}{Disgust}  	& 1	& Inner Brow Raiser & \multirow{3}{*}{\includegraphics[width=0.1\textwidth]{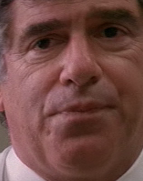}} \\
						   			& 4  &  Brow Lowerer &  \\
                                    & 6  &  Cheek Raiser &  \\
        \\
        \\
		\hline
         \multirow{2}{*}{Fear}  	& 5	& Upper Lid Raiser & \multirow{3}{*}{\includegraphics[width=0.1\textwidth]{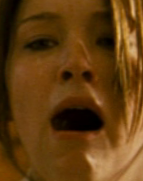}} \\
						   			& 25  &  Lips Part &  \\
        \\
        \\
        \\      
		\hline
        \multirow{3}{*}{Happy}  	& 6  &  Cheek Raiser & \multirow{3}{*}{\includegraphics[width=0.1\textwidth]{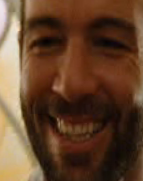}} \\
						   			& 12  &  Lip Corner Puller &  \\
                                    & 25  &  Lips Part &  \\
        \\
        \\      
		\hline
       	 \multirow{2}{*}{Sad}  		& 12  &  Lip Corner Puller & \multirow{2}{*}{\includegraphics[width=0.1\textwidth]{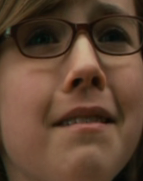}} \\
						   			& 17  &  Chin Raiser &  \\
        \\
        \\
        \\
		\hline
        \multirow{3}{*}{Surpise}  	& 2  &  Outer Brow Raiser & \multirow{3}{*}{\includegraphics[width=0.1\textwidth]{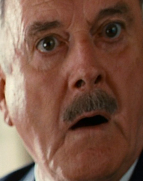}} \\
						   			& 25  &  Lips Part &  \\
                                    & 26  &  Jaw Drop &  \\
        \\
        \\
		\hline
	\end{tabular}
\end{table}

The table shows the coherence of the approach when recognizing the AUs in SFEW images. Given the fact that only 11 AU labels are used in the EmotioNet database (which is the database used to train the SJMT approach), some of the AUs theoretically forming the emotions cannot be recognized. A clear example is the Sad emotion, where the most discriminant AU in theory is 15 (Lip Corner Depressor) and the model recognizes it as 12 (Lip Corner Puller).

\subsection{Compound emotion recognition}

Most of the research on emotion recognition has focused on the study of the seven basic categories. We use the term ``basic" to refer to the fact that such emotion categories cannot be decomposed into smaller semantic labels. However, it has been shown~\cite{du2014compound} that many more facial expressions of emotion exist and are used regularly by humans. Compound emotions are those that can be constructed by combining basic component categories to create new ones. Hence, in this section we will validate our method on the extended problem of recognizing the following compound emotions from the EmotioNet database: \emph{angrily disgusted}, \emph{angrily surprised}, \emph{fearfully angry}, \emph{fearfully surprised}, \emph{happily disgusted}, \emph{happily surprised}, \emph{sadly angry}, and \emph{sadly disgusted}. The number of available samples is scarce for most of the classes, and we hypothesize that this emotion prediction task can take benefit from a joint learning process along with an auxiliary task such as AU recognition.

Following the experiments of the previous sections, here we compare the performance of the SJMT approach for compound emotion images with a single dedicated task network, in both cases training a ResNet 110. Given the unbalanced number of images per class (see Table~\ref{tab:tab_results_compound}), we decided to train the networks with a training set of 160 images consisting of 15 images in each class.

\begin{table}[ht]
	\renewcommand{\arraystretch}{1.3}
	\caption{Results of single and multi-task learning for compound emotion recognition tasks in percentage of accuracy.}
	\label{tab:tab_results_compound}
	\centering
	\begin{tabular}{*4c}	
		\hline
		\multirow{2}{*}{Compound} 	& Number & Single & SJMT \\ 									emotion	&  images & ResNet-110 & ResNet-110 \\
		\hline
		Angrily disgusted & 19 & 15.8\% & 42.1\% \\
        Angrily surprised & 25 & 36.0\% & 64.0\% \\
        Fearfully angry & 19 & 15.8\% & 63.1\% \\
        Fearfully surprised & 17 & 58.8\% & 70.6\% \\
        Happily disgusted & 486 & 47.6\% & 78.6\% \\
        Happily surprised & 36 & 61.1\% & 72.2\% \\
        Sadly angry & 15 & 13.3\% & 20.0\% \\
        Sadly disgusted & 105 & 31.4\% & 64.7\% \\
		\hline
        Mean compounds & - & 34.9\% & 59.4\% \\
        All images & 722 & 43.4\% & 73.1\% \\
        \hline
	\end{tabular}
\end{table}

Table~\ref{tab:tab_results_compound} shows that also when dealing with more complex emotions the SJMT network outperforms the single one, globally and for each of the classes. The improvement on the performance of the classes with the most examples (happily disgusted and sadly disgusted) also explains the large difference between methods when comparing globally, since much more images from the total are correctly classified. However, when comparing the mean of all the class accuracies we can see that SJMT with 59.4\% considerably improves the single network performance (34.9\%).

\section{Conclusions}
\label{sec:conclusions}

In this paper we presented the \emph{Selective Joint Multi-task} approach which defines a novel dataset-wise selective sigmoid cross-entropy loss function to address multi-task, multi-label and multi-database problems. Specifically, we addressed this proposal to overcome one of the challenges with emotion recognition \emph{in the wild}, which is the lack of large public labeled databases by including AU recognition in a multi-task approach. 

Our proposed approach was assessed using the SFEW database for emotion recognition and the EmotioNet database for AU detection, and the results were compared to single task networks dedicated to each task individually and to the classical multi-task approach. SJMT obtained the best results in terms of accuracy for all the experiments. Results also showed the benefits of learning multiple correlated tasks simultaneously by demonstrating visually that even for images without AU labels, the model is capable of inferring them in a coherent way. Finally, we evaluated the proposed method in a database of compound emotions, again obtaining improved results.

Future work will consist of studying and developing the inclusion of the landmark detection task in this multi-task approach. We believe that the addition of structural and geometrical cues to the appearance based CNNs might boost the performance of the emotion recognition task, but a new reformulation of the selective cross-entropy function needs to be done in order to handle such a problem in a joint training.

\begin{acknowledgements}
This research was supported by TIN2015-66951-C2-2-R grant from the Spanish Ministry of Economy and Competitiveness and NVIDIA Hardware grant program.
\end{acknowledgements}

\bibliographystyle{spmpsci}      
\bibliography{bib_file}{}  

%
%

\end{document}